\documentclass[runningheads]{llncs}
\usepackage{booktabs}
\usepackage{multirow}
\usepackage{xcolor}
\usepackage{float}
\definecolor{ForestGreen}{RGB}{34,139,34}
\definecolor{BrickRed}{RGB}{178,34,34}
\usepackage[T1]{fontenc}
\usepackage{graphicx}
\usepackage{booktabs}
\usepackage[misc]{ifsym}
\usepackage{amsmath}
\usepackage{amssymb}
\usepackage{multirow}
\usepackage{enumitem}
\usepackage{float}
\usepackage{adjustbox}
\usepackage{hyperref}
\hypersetup{
    colorlinks=true,
    linkcolor=blue,
    citecolor=blue,
    urlcolor=blue
}

\usepackage{xcolor}
\usepackage{url}
\usepackage{longtable}
\usepackage{microtype}
\usepackage{capt-of}
\usepackage{placeins}

\begin{document}

\title{FIG-MAC: A Fine-grained Inspiration Graph Empowered Multi-Agent Collaboration for Automated Scientific Hypothesis Generation}

\titlerunning{FIG-MAC for ASHG}

\author{Siyuan Chen\inst{1} \and
Yuping Gao\inst{1} \and
Tai Huang\inst{1} \and
Yuling Zhou\inst{1} \and
Dongyu Su\inst{1} \and
Tongguan Wang\inst{1} \and
Ying Sha\inst{1}\textsuperscript{(\Letter)}}

\authorrunning{S. Chen et al.}

\institute{\textsuperscript{1} Huazhong Agricultural University, Wuhan, China\\
\email{shaying@mail.hzau.edu.cn}}

\maketitle

\begingroup
\renewcommand\thefootnote{}
\footnotetext{Siyuan Chen, Yuping Gao, Tai Huang and Yuling Zhou---Equal contribution.}
\addtocounter{footnote}{-1}
\endgroup

\begin{abstract}
Automated Scientific Hypothesis Generation (ASHG) leveraging large language models (LLMs) aims to harness their strengths in knowledge integration, logical reasoning, and text generation for efficiently producing innovative, verifiable scientific hypotheses. However, existing approaches face three critical limitations: (i) insufficient fine-grained knowledge representation, as unstructured context impairs LLM comprehension and reasoning; (ii) lack of knowledge evolution traceability, constraining the generation of innovative cross-domain conjectures; (iii) absence of human-inspired and role-specialized multi-agent collaboration, failing to leverage the efficient innovation mechanisms arising from cognitive diversity and division of labor among expert roles to complete the full ASHG-evaluation loop. To address these limitations, we propose the Fine-grained Inspiration Graph-empowered Multi-Agent Collaboration framework (FIG-MAC) including: (i) Fine-grained Inspiration Graph module (FIG), to obtain fine-grained structured knowledge representations, where nodes represent academic semantic units and edges encode heuristic relationships among them; (ii) ontology-guided hybrid reasoning module, which follows the FIG ontology to construct structured reasoning paths and uses vector search to enrich them with semantic evidence, to obtain the traceable scientific conjecture paths; (iii) role-specialized Multi-Agent framework, inspired by real scientific teamwork, enabling a closed-loop process of scientific conjecture generation, screening, and validation. 
Experimental results demonstrate that FIG-MAC significantly outperforms state-of-the-art methods on source diversity, provenance-adjusted novelty, and raw novelty.

\end{abstract}

\section{Introduction}
\begin{figure}[t]
    \centering 
    \includegraphics[width=0.98\textwidth]{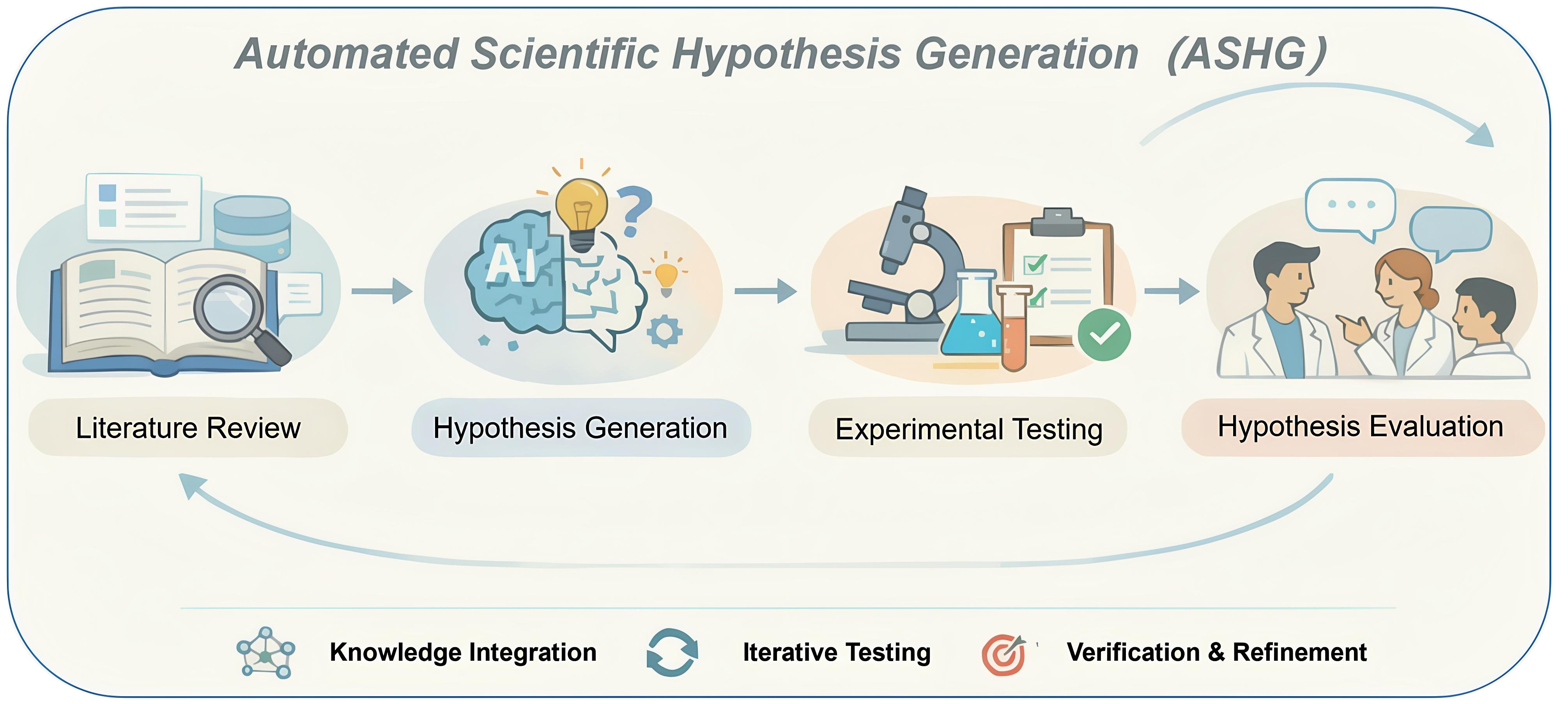} 
    \caption{The ASHG pipeline: \textit{Literature Review} $\rightarrow$ \textit{Hypothesis Generation} $\rightarrow$ \textit{Experimental Testing} $\rightarrow$ \textit{Hypothesis Evaluation}, with iterative feedback loops. Core mechanisms include Knowledge Integration, Iterative Testing, Verification and Refinement.}
    \label{fig1}
\end{figure}

Automated Scientific Hypothesis Generation (ASHG) based on large language models (LLMs) aims to leverage their capabilities in knowledge integration, reasoning, and generation to produce innovative and verifiable scientific hypotheses~\cite{agarwal2025litllmsllmsliteraturereview}. As illustrated in Figure~\ref{fig1}, such methods have shown early success in autonomous chemical research~\cite{boiko2023autonomous}.

In recent years, ASHG focus on three domains: (1) to identify potential research gaps in academic literature, models such as TrustResearcher~\cite{DBLP:journals/corr/abs-2510-20844}, HypoGen~\cite{zhou2024hypothesis} detect research blind spots by integrating knowledge retrieval and logical reasoning; (2) to improve the empirical feasibility and verifiability of generated hypotheses, AI Scientist systems~\cite{aiscientist_v2} operationalize hypothesis generation as an iterative “proposing-evaluating-refining” loop, Causal Graph-guided approaches~\cite{tong2024automating} constrain hypothesis space using explicit causal structures, and LLM-based iterative optimization~\cite{rabby2025iterative} progressively refines hypotheses via feedback-driven reasoning and evaluation; (3) to improve the factual consistency and verifiability of hypotheses, techniques including Retrieval-Augmented Generation (RAG) and Graph Reasoning~\cite{liang2024survey} are used to enhance the scientific rigor and reliability of hypotheses.

Despite this progress, three critical limitations remain: (i) Insufficient fine-grained knowledge representation: existing external knowledge is typically abstract-level, producing coarse-grained representations that lack complete logical links among scientific concepts, thereby impairing LLM reasoning accuracy; (ii) Lack of traceability in knowledge evolution: current methods reason over surface-level semantic similarity or static citation links, failing to model the origin and evolution path of knowledge and thus constraining cross-domain hypothesis generation; (iii) Absence of role-specialized multi-agent collaboration: existing systems rarely replicate the cognitive diversity and division of labor found in human research teams, making it difficult to close the full ASHG-evaluation loop.

%补充语义单元解释
To address these challenges, we propose the Fine-grained Inspiration Graph-empowered Multi-Agent Collaboration framework (FIG-MAC). First, the Fine-grained Inspiration Graph (FIG) module transforms raw corpus into structured semantic units (abstract, core problem, related work, preliminary innovation analysis, research questions and corresponding solutions proposed in the paper, and framework summary, etc.) through knowledge parsing and constructs relationships between these units, providing structured knowledge for reasoning.
Second, the ontology-guided hybrid reasoning module generates structured knowledge evolution paths via graph traversal, and supplements each reasoning node with domain-specific semantic evidence through vector retrieval, forming a traceable dynamic knowledge context. Finally, the role-specialized Multi-Agent framework simulates the real scientific research process by setting intelligent roles such as knowledge exploration, idea generation, and parallel domain review. Through multiple rounds of evaluation and knowledge interpretability feedback, the framework generates rigorously verified scientific hypotheses. Extensive experiments demonstrate that FIG-MAC significantly outperforms state-of-the-art methods: improving source diversity by 26.9\%, provenance-adjusted novelty by 55.1\%, and overall novelty by 31.8\% compared to the best baseline.

Our contributions are summarized as follows:

\begin{itemize}[leftmargin=*,itemsep=1pt,parsep=0pt,topsep=2pt,partopsep=0pt]
\item \textbf{Fine-grained Inspiration Graph (FIG).} We construct a heterogeneous knowledge graph that decomposes papers into fine-grained semantic units and introduces Inspired edges to explicitly model cross-paper innovation pathways.
\item \textbf{Ontology-Guided Hybrid Reasoning.} We propose a ontology-guided hybrid reasoning mechanism where graph traversal follows the FIG ontology to discover traceable cross-domain knowledge evolution paths, while vector retrieval enriches these paths with domain-specific semantic evidence. This hybrid approach ensures generated hypotheses achieve both structural novelty through cross-domain connections and technical feasibility through grounded semantic evidence.
\item \textbf{Multi-Agent Collaboration Framework for ASHG.} We design a role-specialized multi-agent system that simulates real scientific teamwork through a closed-loop process where generation, evaluation, and validation are tightly coupled, ensuring hypotheses are both innovative and scientifically rigorous.
\end{itemize}

\section{Related Work}
%按照科学猜想流程介绍

% We structure related work along the scientific hypothesis discovery pipeline: evidence retrieval, hypothesis generation, and collaborative validation.

\subsection{Evidence Retrieval in Scientific Literatures}
Scientific paper retrieval employs queries like abstracts/titles \cite{cohan2020specter}, keywords \cite{raamkumar2017keywords} aspects, relying on semantic similarity for representation learning. However, mainstream retrieval models (sentence-based similarity \cite{reimers2019sentence} or generic inter-document embedding \cite{cohan2020specter}) often yield irrelevant results, emphasizing the need for deeper analysis of solution applicability to target problems.

\subsection{From Literature-Based Discovery to LLM-Based Generation}
Automated scientific hypothesis generation has evolved from traditional statistical methods to modern deep learning approaches. Literature-Based Discovery (LBD) is rooted in "undiscovered public knowledge" and the ABC model infers relationships between concepts $A$ and $C$ via shared intermediate concepts $B$~\cite{swanson1986fish}. Recent LLM-based approaches enhance hypothesis generation through agentic frameworks~\cite{DBLP:journals/corr/abs-2510-20844}, but may still suffer from retrieval bias and evidence selection issues, limiting hypothesis novelty, validity and diversity.

% Keep this sparsely populated page naturally ragged rather than stretching subsection spacing.
\thisbottomragged
\subsection{Structured Knowledge for Collaborative Validation}
Knowledge graph-based approaches frame hypothesis generation as link prediction over structured scientific knowledge \cite{borrego2025research}. Early academic knowledge graphs \cite{ammar2018construction} model coarse-grained entities (papers, authors, venues) and citations, while recent work focuses on fine-grained entity/relationship modeling \cite{luan2018multi} for precise hypothesis generation.

Multi-agent systems enable collaborative hypothesis generation/validation by distributing roles (generation, critique, refinement) across specialized agents \cite{ghafarollahi2025sciagents}. SciAgents integrates ontological knowledge graphs with LLMs via specialized agents (Ontologist, Scientists, Critic) \cite{ghafarollahi2025sciagents}, and frameworks like POPPER implement agentic validation based on falsification \cite{huang2025automated}.

Despite these progress, existing methods still face key limitations: (1) reliance on shallow semantic similarity or static graph links that ignore dynamic knowledge evolution and closed-loop evaluation; (2) lack of unified frameworks for discovery, traceable reasoning, and multi-agent evaluation; and (3) separation of hypothesis generation and validation, which hinders iterative refinement through empirical feedback. These limitations motivate our framework, which combines fine-grained knowledge construction, traceable reasoning paths, and role-specialized multi-agent collaboration for hypothesis generation and validation.

\section{Methodology}
\label{sec:methodology}

\begin{figure}[t]
\centering
\includegraphics[width=1\textwidth]{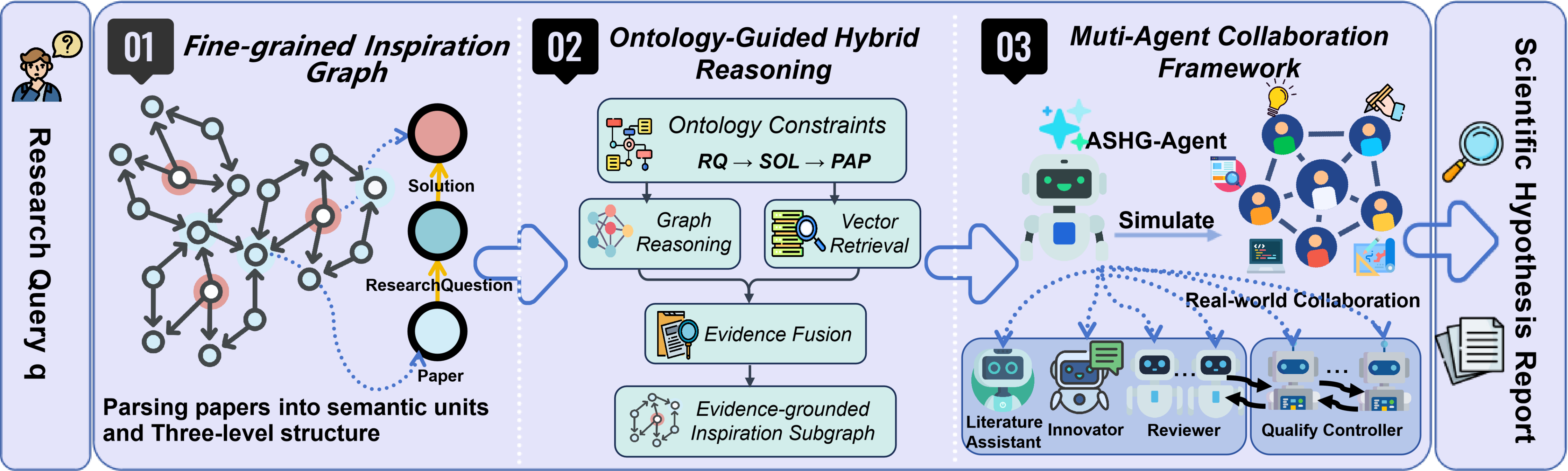} 
\caption{Overview of the FIG-MAC framework:
(\textbf{1}) \textit{Fine-grained Inspiration Graph (FIG)}: academic papers are parsed into semantic units and organized into a three-level PAP–RQ–SOL structure connected by hierarchical and cross-paper inspiration relations.
(\textbf{2}) \textit{Ontology-Guided Hybrid Reasoning}: graph-based FIG traversal follows ontology constraints to discover structured inspiration paths, which are enriched by vector retrieval with domain-specific semantic evidence.
(\textbf{3}) \textit{Multi-Agent Collaboration Framework}: role-specialized agents comprising a Literature Review Assistant, Innovator, Reviewer, and Quality Controller collaboratively produce a Scientific Hypothesis Report.}
\label{fig2}
\end{figure}

FIG-MAC is a collaborative multi-agent framework designed for ASHG by simulating the reasoning process of human research teams. As shown in Figure~\ref{fig2}, it comprises three components: (i) the Fine-grained Inspiration Graph module (FIG); (ii) the ontology-guided hybrid reasoning module; and (iii) the role-specialized Multi-Agent framework.

\subsection{Fine-grained Inspiration Graph Construction}
\label{sec:graph_construction}

To enable structured knowledge representation that captures fine-grained academic semantics and cross-paper innovation pathways, we design the Fine-grained Inspiration Graph (FIG)\footnote{\url{https://github.com/neo4j/neo4j}}. FIG decomposes scientific literature into semantic units and models their relationships at multiple granularities. Formally, FIG is a heterogeneous knowledge graph denoted as $\mathcal{G} = (\mathcal{V}, \mathcal{E}, \mathcal{R})$, where $\mathcal{V}$ represents the set of nodes, $\mathcal{E}$ denotes the set of edges, and $\mathcal{R}$ is the set of relation types. The current implementation yields three primary entity types: $PAP$ (Papers), $RQ$ (Research Questions), and $SOL$ (Solutions).

\textbf{Fine-grained Data Structuring.} 
The core challenge lies in accurately extracting key information and entities from raw text. 
We perform fine-grained structured parsing on each paper, dividing the full text into 21 semantic units. 
These include \textit{metadata attributes} (DOI, title, authors, year, conference, venue, citation count, and abstract) that provide bibliographic context; 
\textit{core content attributes} capturing the main knowledge elements, including the core problem, related work, preliminary innovation analysis, up to four research questions (RQ1–RQ4), up to four solutions (SOL1–SOL4), and a framework summary; 
and \textit{auxiliary attributes} that provide additional semantic annotations for entity disambiguation and relationship inference. 
Together, these components form a three-level knowledge structure centered on the core entities \textit{PAP–RQ–SOL}.

\textbf{Entity Extraction and Relation Construction.} Papers are manually annotated and refined to standardized samples (Kappa=0.85). And we fine-tune a Qwen-based llm as a extractor. This model then processes the full corpus, generating structured semantic units for each paper. FIG encodes two relation classes. Hierarchical relations $r_{\text{has\_rq}}$ ($PAP_i$ to $RQ_j$) and $r_{\text{has\_sol}}$ ($RQ$ to $SOL$) map intra-paper structure. Cross-paper relations $r_{\text{related}}$ (semantically similar entities across papers) and $r_{\text{inspired}}$ capture inter-paper innovation pathways.

To construct cross-paper edges at scale, we combine similarity filtering with supervised classification. Candidate pairs are manually labeled by three experts into three categories: Related, Inspired, or Irrelevant; below a similarity threshold are pruned; the remaining candidates are classified by a fine-tuned Qwen3-8B model\footnote{\url{https://github.com/QwenLM/Qwen3}}. An inspired edge $r_{\text{inspired}}$: $SOL_i$ to $RQ_j$ denotes that the solution proposed in paper $Pi$ serves as a conceptual or methodological precursor to the research question posed in paper $Pj$, where $Pi$ and $Pj$ need not share a citation relationship. This edge explicitly models latent knowledge transfer paths that are invisible to citation-based analysis.

To discover invisible inspiration paths, we employ a Relation Graph Convolutional Network (RGCN)\footnote{\url{https://github.com/awslabs/graphstorm}} for representation learning, where each node aggregates messages from its typed neighbors via relation-specific weight matrices across $l$ layers, yielding embeddings $\mathbf{h}_i^{(l)}$ that capture both local structure and relational context. Link prediction over the learned embeddings then uses the DistMult scoring function, which models each candidate inspired edge $(u, r, v)$ as a bilinear interaction:

\begin{equation}
    s(u, r, v) = (\mathbf{h}_u^L)^T \mathbf{R}_r \mathbf{h}_v^L,
\end{equation}
where $\mathbf{R}_r$ is a learnable diagonal matrix associated with relation $r$, higher scores indicate stronger evidence for the inspired relation.

\subsection{Ontology-Guided Hybrid Reasoning Module}
\label{sec:hybrid_reasoning}

We propose a ontology-guided hybrid reasoning method that integrates two complementary paradigms: graph-based path discovery follows the FIG ontology to construct structured paths of cross-domain knowledge evolution, while vector-based semantic search enriches these paths with domain-specific technical evidence. Given a research query $Q$, two retrieval functions operate in parallel: (1) VectorEnrichment $\mathcal{R}_V(Q)$ identifies topically relevant papers via dense embeddings, returning papers ranked by semantic similarity; (2) \text{SubgraphSkeleton} $\mathcal{R}_G(Q)$ traverses FIG to extract inspiration chains $\pi$ connecting $Q$ to $\text{SOL}$ nodes through $r_{\text{inspired}}$ edges. The integration function $\mathcal{I}$ merges structural backbones with semantic evidence:
\begin{equation}
    \mathcal{I}(\mathcal{R}_V, \mathcal{R}_G) = \text{SubgraphSkeleton}(\mathcal{R}_G) \cup \text{VectorEnrichment}(\mathcal{R}_V),
\end{equation}
$\text{SubgraphSkeleton}(\mathcal{R}_G)$ extracts multi-hop inspiration chains from FIG, capturing structural novelty through cross-domain connections; $\text{VectorEnrichment}(\mathcal{R}_V)$ retrieves semantically similar papers, ensuring technical grounding. Their union forms an inspiration subgraph that is both structurally diverse and semantically coherent. Each inspiration chain $\pi = (\text{RQ}_0, \text{RQ}_1, \ldots, \text{RQ}_m, \text{SOL}_j, \text{PAP}_k)$ is ranked by:
\begin{equation}
    \text{score}(\pi) = \alpha \cdot \text{sim}(Q, \text{RQ}_0) + \beta \cdot \text{conf}(\text{RQ}_m \rightarrow \text{SOL}_j) + \gamma \cdot \text{conf}(\text{SOL}_j \rightarrow \text{PAP}_k),
    \label{eq:score}
\end{equation}
The three terms balance complementary quality signals. $\text{sim}(Q,\text{RQ}_0)$ is the cosine similarity between the query and the path entry node, ensuring topical relevance. $\text{conf}(\text{RQ}_m \rightarrow \text{SOL}_j)$ is the RGCN-predicted score for the resolution link from the final intermediate question to its solution. $\text{conf}(\text{SOL}_j \rightarrow \text{PAP}_k)$ measures the inspired-relation strength from that solution to the target paper. The weights $\alpha$, $\beta$, $\gamma$ are tuned on the validation set. Beam search extracts multiple high-scoring paths to form the final inspiration chain set.

\subsection{Role-Specialized Multi-Agent Framework}
\label{sec:workflow}

To enable collaborative hypothesis generation that mimics the coordination and brainstorming dynamics of real research teams, we design a Multi-Agent Collaboration framework with multiple specialized agent roles. Each agent maintains a role-specific policy over its action space, an observation function over its current state, a dual-memory structure combining persistent long-term knowledge with short-term conversational history. State transitions follow a Markov decision process in which agents observe their current state, sample actions from their role-specific policy, and update their state based on shared context aggregated from retrieval evidence and inter-agent messages.

\textbf{Agent Roles.} The framework comprises seven functional role types covering the full scientific workflow: (1)~\textit{Research Assistants} for literature gathering and synthesis; (2)~\textit{Innovators} for hypothesis drafting; (3)~\textit{Technical Reviewers} for scientific validity assessment; (4)~\textit{Practical Reviewers} for feasibility evaluation; (5)~\textit{Ethical Reviewers} for societal impact analysis; (6)~\textit{Coordinators} for perspective integration; and (7)~\textit{Quality Controllers} for iterative refinement. Each role type may be instantiated by one or more agents depending on workload and parallelism requirements.

\textbf{Parallel Review.} During the analysis phase, three independent reviewers (Technical, Practical, Ethical) operate concurrently, each producing a review vector across five dimensions: Novelty, Significance, Effectiveness, Clarity, and Feasibility. The three review vectors are aggregated via confidence-aware softmax weighting, where higher-confidence reviewers receive greater weight and a temperature parameter $\tau$ controls sharpness.

\textbf{State-Driven Workflow.} The collaboration follows a finite state machine with seven phases. After the review phase, the system transitions to POLISH when the aggregate review score meets the predefined quality threshold or the iteration budget is reached; otherwise it enters REVISION for further refinement. During revision, Quality Controller agents update the hypothesis based on structured reviewer feedback, mirroring the peer-review process and enabling progressively refined hypotheses with traceable revisions.

\section{Experiments}
\subsection{Evaluation Setup}

\textbf{Datasets.} We evaluate FIG-MAC on the Fine-grained Paper Dataset (FPD), constructed from five premier conferences covering AI research from 2019 to 2024 with fine-grained semantic annotations, as shown in Table~\ref{tab:dataset_sources}. Each paper is parsed into 21 semantic units forming a three-level PAP - RQ - SOL structure. We randomly selected 150 diverse research questions (RQs) spanning multiple subfields and partitioned them into a validation set (50 RQs) for hyperparameter tuning and a held-out test set (100 RQs) for final evaluation. All reported objective metrics, statistical tests, and human evaluation use the held-out test set only.

\begin{table}[t]
\centering
\caption{Dataset Paper Sources by Conference.}
\label{tab:dataset_sources}
\small
\begin{tabular}{lcccccc}
\toprule
\textbf{Conference} & ACL & EMNLP & NAACL & EACL & AAAI & \textbf{Total} \\
\midrule
\textbf{Paper Counts} & 5,877 & 7,539 & 2,086 & 991 & 10,424 & 26,917 \\
\bottomrule
\end{tabular}
\end{table}

\textbf{Baseline.} We compare our framework with three state-of-the-art methods: (1) Virtual-Scientists (VirSci)~\cite{su2025many}, a multi-agent method with team-based discussion mechanisms; (2) AI Scientist-v2~\cite{DBLP:journals/corr/abs-2408-06292}, an end-to-end agentic method using tree search for discovery; (3) CoI-Agent~\cite{li2024chain}, a framework simulating a chain of ideas for research development.

\textbf{Evaluation Metrics.}
We jointly considering objective and subjective metrics following the established evaluation methodology in prior work~\cite{su2025many}. For objective novelty, we retrieve the top-$K=10$ semantically similar papers for each hypothesis and partition them into \textit{historical papers} $\mathcal{P}_{\text{past}}$ (before 2022) and \textit{contemporary papers} $\mathcal{P}_{\text{contemp}}$ (2022 or later). Three indicators are computed: (1) Historical Dissimilarity (HD), measuring the semantic distance from prior literature~\cite{shao2020bertpli,DBLP:conf/aaai/ZhouSGTSLXJ24}; Contemporary Dissimilarity (CD), capturing alignment with recent work~\cite{wu2019large}; and Contemporary Impact (CI), estimating significance via citation percentile ranks of related contemporary papers~\cite{yang2022gender}. These are integrated into Overall Novelty:
\begin{equation}
    \text{ON}_{\text{raw}}(h) = \frac{\text{HD}(h) \times \text{CI}(h)}{\text{CD}(h)}
\end{equation}
where higher values indicate hypotheses that deviate from historical work, connect to high-impact research topics, and remain aligned with contemporary developments.

For provenance-adjusted novelty, we compute source similarity $S_{\text{src}}$ (average cosine similarity between the hypothesis and its retrieved sources) and source diversity $U_{\text{src}}$ (average pairwise dissimilarity among sources, with higher values indicating cross-domain coverage). These combine into a provenance factor $G(h)=\alpha_2(1-S_{\text{src}}(h))+\beta_2U_{\text{src}}(h)$, which modulates the raw novelty estimate:

\begin{equation}
P(h)=\text{ON}_{\text{raw}}(h)\left[\gamma_2G(h)+(1-\gamma_2)\right].
\label{eq:p_score}
\end{equation}

Subjective quality is assessed by Qwen-Max on five dimensions (Novelty, Significance, Effectiveness, Clarity, Feasibility) on a 1-10 scale.

\textbf{Experimental Configuration.} We use a FAISS vector database containing 26,917 papers and a knowledge graph comprising 26,917 PAP, 76,933 RQ, and 76,933 SOL nodes. All methods use Qwen text-embedding-v2 with  $d=1536$ for semantic similarity. The RGCN uses a 2-layer architecture with hidden dimension 256 and DistMult decoder, trained for 30 epochs with early stopping, achieving validation MRR of 0.4563 and test MRR of 0.4599 on inspired edge prediction. All experiments are conducted on NVIDIA RTX 4090 GPUs  using PyTorch 2.0 with DGL 1.1.

\subsection{Main Results}
\label{sec:main_results}

Table~\ref{tab:main_results} presents comprehensive evaluation across all methods. FIG-MAC achieves the highest source diversity ($U_{\text{src}} = 0.650$) and competitive provenance-adjusted novelty ($P=0.535$), demonstrating superior cross-domain knowledge integration. The raw novelty $ON_{\text{raw}}=0.684$ confirms the effectiveness of graph-enhanced hybrid reasoning. Most notably, FIG-MAC achieves the lowest CD, among all methods. Our primary experiments employ Qwen-Max as the backbone LLM. To assess generalization across different backbones, we extend evaluation to two additional open-source models. Table~\ref{tab:cross_model} reports cross-model objective metrics.

\begin{table}[t]
\centering
\caption{Main results on the held-out test set (100 RQs). \textbf{Left:} Objective metrics (higher is better except CD, $S_{\text{src}}$) for semantic novelty (HD, CD), provenance ($G$, $P$), and diversity (CI, $U_{\text{src}}$). \textbf{Right:} Expert ratings (1--10) for novelty (Nov.), significance (Sig.), effectiveness (Eff.), clarity (Cla.), and feasibility (Fea.). AI Scientist is non-RAG.}
\label{tab:main_results}
\label{tab:subjective_quality}
\small
\setlength{\tabcolsep}{2pt}
\renewcommand{\arraystretch}{0.95}
\begin{adjustbox}{max width=\textwidth}
\begin{tabular}{lccccccccccccc}
\toprule
\multirow{2}{*}{\textbf{Method}} 
& \multicolumn{8}{c}{\textbf{Objective metrics}} 
& \multicolumn{5}{c}{\textbf{Subjective quality (1--10)}} \\
\cmidrule(lr){2-9} \cmidrule(lr){10-14}
& \textbf{HD} & \textbf{CD} & \textbf{CI} & \textbf{ON$_{\text{raw}}$} & \textbf{S$_{\text{src}}$} & \textbf{U$_{\text{src}}$} & \textbf{G} & \textbf{P} 
& \textbf{Nov.} & \textbf{Sig.} & \textbf{Eff.} & \textbf{Cla.} & \textbf{Fea.} \\
\midrule
Virtual Scientists & 0.468 & \textbf{0.370} & 0.506 & 0.504 & 0.580 & 0.260 & 0.340 & 0.271 & 8.5 & 7.6 & 7.0 & 8.5 & 7.2 \\
COI Agent          & 0.477 & 0.375 & \textbf{0.517} & 0.519 & 0.473 & 0.512 & 0.520 & 0.345 & 8.4 & 8.2 & 7.0 & 8.1 & 7.0 \\
AI Scientist       & \textbf{0.503} & 0.389 & 0.490 & 0.504 & N/A  & N/A  & N/A  & N/A  & 8.0 & 7.2 & 7.6 & 7.3 & 7.0 \\
\midrule
\textbf{FIG-MAC (Ours)} 
& 0.500 & 0.291 & 0.535 & \textbf{0.684} & 0.276 & \textbf{0.650} & \textbf{0.687} & \textbf{0.535} 
& \textbf{8.7} & \textbf{8.4} & \textbf{7.8} & \textbf{8.6} & \textbf{7.5} \\
\bottomrule
\end{tabular}
\end{adjustbox}
\end{table}

\textbf{Human Evaluation.}
To validate the LLM-based subjective assessment, we conducted a blinded evaluation on the held-out 100 test RQs. A pool of five CS PhD students evaluated the same five dimensions (Novelty, Significance, Effectiveness, Clarity, and Feasibility); each RQ was independently scored by two students, yielding 200 assessments. A third reviewer adjudicated cases in which the first two scores differed by more than two points on any dimension (approximately 33.3\% of cases). Reports were presented in randomized order without method identifiers. Before formal scoring, annotators completed a calibration session on 10 pilot hypotheses using the same structured 1--10 rubric as the LLM judge, with explicit anchors for each dimension. As summarized in Table~\ref{tab:human_eval_stats}, human ratings (mean 7.32) are systematically lower than LLM ratings (mean 7.78), with agreement within $\pm1$ point ranging from 86.0\% to 100.0\% across dimensions and significant rank correlation ($p < 0.05$) for four of five dimensions. This gap is consistent with known leniency bias in LLM-as-Judge settings, but the rank ordering of methods is preserved, supporting the validity of our LLM-based comparisons. The 50 validation RQs were used only for hyperparameter tuning and were not included in the human evaluation.

\textbf{Statistical Significance.}
We performed paired Wilcoxon signed-rank tests across the held-out 100 test RQs under identical settings; the 50 validation RQs were excluded from all final statistical comparisons. The Shapiro-Wilk test confirmed non-normal distributions of differences ($p < 0.05$), justifying the non-parametric choice. We report Cohen's $d = \bar{\Delta}/s_{\Delta}$ for paired samples to quantify practical significance. All comparisons are statistically significant ($p < 0.001$) with medium-to-very-large effect sizes (d=0.79--2.77). After Bonferroni correction and Holm adjustment, all $p$-values remain significant, and the design achieves $>0.9$ statistical power for medium effects. These gains in source diversity and provenance-adjusted novelty are both statistically robust and practically meaningful.

\begin{table}[H]
\centering

\begin{minipage}[t]{0.48\textwidth}
\centering
\caption{Human evaluation on the held-out 100 test RQs (N=200 assessments, 5 CS PhD students). $\rho$: Spearman rank correlation; Agree$_{\pm1}$: percent agreement within $\pm$1 point.}
\label{tab:human_eval_stats}
\scriptsize
\setlength{\tabcolsep}{2pt}
\renewcommand{\arraystretch}{0.95}
\begin{adjustbox}{max width=\linewidth}
\begin{tabular}{lccccc}
\toprule
\textbf{Dim.} & \textbf{Mean} & \textbf{Std} & $\boldsymbol{\rho}$ & \textbf{p} & \textbf{Agree$_{\pm1}$} \\
\midrule
Novelty       & 7.03 & 0.55 & 0.419 & $<$.01  & 98.0\% \\
Significance  & 7.65 & 0.56 & 0.356 & $<$.05  & 100.0\% \\
Effectiveness & 7.08 & 0.90 & 0.494 & $<$.001 & 86.0\% \\
Clarity       & 7.92 & 0.56 & 0.198 & .169    & 96.0\% \\
Feasibility   & 6.91 & 0.85 & 0.456 & $<$.001 & 86.0\% \\
\midrule
Overall       & 7.32 & 0.46 & -- & -- & -- \\
\bottomrule
\end{tabular}
\end{adjustbox}
\end{minipage}
\hfill
\begin{minipage}[t]{0.48\textwidth}
\centering
\caption{Cross-model generalization. AI Scientist is non-RAG, so $P$ and $U_{\text{src}}$ are N/A.}
\label{tab:cross_model}
\scriptsize
\setlength{\tabcolsep}{2.2pt}
\renewcommand{\arraystretch}{0.95}
\begin{adjustbox}{max width=\linewidth}
\begin{tabular}{lcccc}
\toprule
\textbf{Backbone LLM} & \textbf{Method} & \textbf{ON$_{\text{raw}}$} & \textbf{P} & \textbf{U$_{\text{src}}$} \\
\midrule
\multirow{4}{*}{\textit{Mixtral-8x7b}} 
 & Virtual Scientists & 0.438 & 0.318 & 0.275 \\
 & COI Agent & 0.462 & 0.348 & 0.305 \\
 & AI Scientist & 0.485 & N/A & N/A \\
 & FIG-MAC & \textbf{0.585} & \textbf{0.462} & \textbf{0.565} \\
\midrule
\multirow{4}{*}{\textit{LLaMA3.1-70b}} 
 & Virtual Scientists & 0.508 & 0.372 & 0.342 \\
 & COI Agent & 0.538 & 0.408 & 0.375 \\
 & AI Scientist & 0.562 & N/A & N/A \\
 & FIG-MAC & \textbf{0.622} & \textbf{0.488} & \textbf{0.595} \\
\midrule
\multirow{4}{*}{\textit{Qwen-Max}$^*$} 
 & Virtual Scientists & 0.504 & 0.271 & 0.260 \\
 & COI Agent & 0.519 & 0.345 & 0.512 \\
 & AI Scientist & 0.504 & N/A & N/A \\
 & \textbf{FIG-MAC} & \textbf{0.684} & \textbf{0.535} & \textbf{0.650} \\
\bottomrule
\end{tabular}
\end{adjustbox}
\end{minipage}

\end{table}
\vspace{-12pt}

% \begin{table}[t]
% \centering
% \caption{Cross-model generalization. AI Scientist is non-RAG, so $P$ and $U_{\text{src}}$ are N/A.}
% \label{tab:cross_model}
% \scriptsize
% \setlength{\tabcolsep}{2.2pt}
% \renewcommand{\arraystretch}{0.95}
% \begin{adjustbox}{max width=\textwidth}
% \begin{tabular}{lcccc}
% \toprule
% \textbf{Backbone LLM} & \textbf{Method} & \textbf{ON$_{\text{raw}}$} & \textbf{P} & \textbf{U$_{\text{src}}$} \\
% \midrule
% \multirow{4}{*}{\textit{Mixtral-8x7b}} 
%  & Virtual Scientists & 0.438 & 0.318 & 0.275 \\
%  & COI Agent & 0.462 & 0.348 & 0.305 \\
%  & AI Scientist & 0.485 & N/A & N/A \\
%  & FIG-MAC & \textbf{0.585} & \textbf{0.462} & \textbf{0.565} \\
% \midrule
% \multirow{4}{*}{\textit{LLaMA3.1-70b}} 
%  & Virtual Scientists & 0.508 & 0.372 & 0.342 \\
%  & COI Agent & 0.538 & 0.408 & 0.375 \\
%  & AI Scientist & 0.562 & N/A & N/A \\
%  & FIG-MAC & \textbf{0.622} & \textbf{0.488} & \textbf{0.595} \\
% \midrule
% \multirow{4}{*}{\textit{Qwen-Max}$^*$} 
%  & Virtual Scientists & 0.504 & 0.271 & 0.260 \\
%  & COI Agent & 0.519 & 0.345 & 0.512 \\
%  & AI Scientist & 0.504 & N/A & N/A \\
%  & \textbf{FIG-MAC} & \textbf{0.684} & \textbf{0.535} & \textbf{0.650} \\
% \bottomrule
% \end{tabular}
% \end{adjustbox}
% \end{table}

\subsection{Ablation Study}
\label{sec:ablation}
\begin{figure}[ht]
\centering
\begin{minipage}[t]{0.5\textwidth}
\centering
\captionof{table}{Ablation on the test set (100 RQs). Hybrid mode consistently performs best; MAC denotes the multi-agent framework.}
\label{tab:ablation_results}
\scriptsize
\setlength{\tabcolsep}{1.5pt}
\renewcommand{\arraystretch}{0.92}
\begin{adjustbox}{max width=\linewidth}
\begin{tabular}{lllccc}
\toprule
\textbf{Backbone} & \textbf{Architecture} & \textbf{Mode} & \textbf{ON$_{\text{raw}}$} & \textbf{P} & \textbf{U$_{\text{src}}$} \\
\midrule
\multirow{8}{*}{\textit{Mixtral-8x7b}}
 & \multirow{4}{*}{Single LLM} & No External Knowledge & 0.175 & 0.078 & N/A \\
 &  & Graph Reasoning & 0.338 & 0.188 & 0.172 \\
 &  & Vector Retrieval & 0.308 & 0.168 & 0.122 \\
 &  & Hybrid Reasoning & 0.378 & 0.262 & 0.328 \\
\cmidrule{2-6}
 & \multirow{4}{*}{MAC} & No External Knowledge & 0.232 & 0.102 & N/A \\
 &  & Graph Reasoning & 0.432 & 0.258 & 0.248 \\
 &  & Vector Retrieval & 0.352 & 0.208 & 0.152 \\
 &  & Hybrid Reasoning & 0.585 & 0.462 & 0.565 \\
\midrule
\multirow{8}{*}{\textit{LLaMA3.1-70b}}
 & \multirow{4}{*}{Single LLM} & No External Knowledge & 0.188 & 0.082 & N/A \\
 &  & Graph Reasoning & 0.358 & 0.198 & 0.185 \\
 &  & Vector Retrieval & 0.328 & 0.178 & 0.132 \\
 &  & Hybrid Reasoning & 0.398 & 0.278 & 0.352 \\
\cmidrule{2-6}
 & \multirow{4}{*}{MAC} & No External Knowledge & 0.248 & 0.108 & N/A \\
 &  & Graph Reasoning & 0.458 & 0.268 & 0.268 \\
 &  & Vector Retrieval & 0.372 & 0.218 & 0.162 \\
 &  & Hybrid Reasoning & 0.622 & 0.488 & 0.595 \\
\midrule
\multirow{8}{*}{\textit{Qwen-Max}$^*$}
 & \multirow{4}{*}{Single LLM} & No External Knowledge & 0.215 & 0.095 & N/A \\
 &  & Graph Reasoning & 0.410 & 0.225 & 0.210 \\
 &  & Vector Retrieval & 0.385 & 0.268 & 0.156 \\
 &  & Hybrid Reasoning & 0.452 & 0.312 & 0.385 \\
\cmidrule{2-6}
 & \multirow{4}{*}{MAC} & No External Knowledge & 0.289 & 0.126 & N/A \\
 &  & Graph Reasoning & 0.512 & 0.301 & 0.298 \\
 &  & Vector Retrieval & 0.421 & 0.245 & 0.182 \\
 &  & \textbf{Hybrid Reasoning} & \textbf{0.684} & \textbf{0.535} & \textbf{0.650} \\
\bottomrule
\end{tabular}
\end{adjustbox}
\end{minipage}
\hfill
\begin{minipage}[t]{0.48\textwidth}
\centering
\vspace{2pt}
\includegraphics[width=\linewidth]{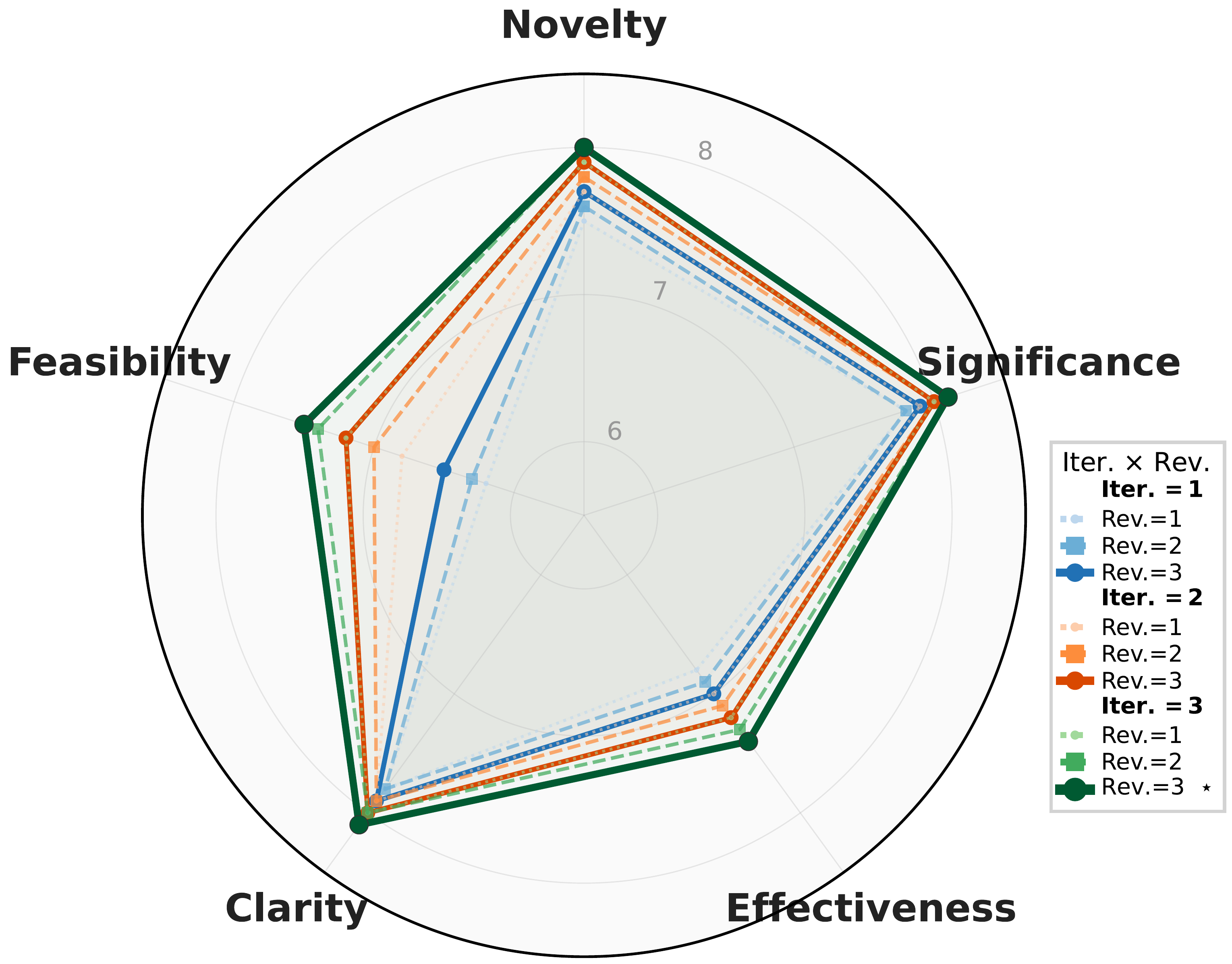}
\caption{Hyperparameter grid search. Radar chart shows subjective scores across all 3$\times$3 (Iter., Rev.) combinations. $\star$ marks the best configuration (Iter.=3, Rev.=3).}
\label{fig:hyperparam_results}
\end{minipage}
\end{figure}

To verify the contribution of each component in FIG-MAC, we conduct a systematic ablation study across two dimensions: (1) reasoning strategy, comparing Hybrid Reasoning, Vector Retrieval, Graph Reasoning, and No External Knowledge; and (2) framework architecture, comparing our multi-agent collaboration (MAC) framework with a Single LLM. Results are summarized in Table~\ref{tab:ablation_results}.

Removing FIG traversal causes $U_{\text{src}}$ to drop sharply, confirming structural reasoning is essential for cross-domain discovery. Our MAC framework consistently outperforms Single LLM, especially in Hybrid Reasoning mode, proving the effectiveness of specialized agents in synthesizing complex evidence.

\subsection{Hyperparameter Analysis}
\label{sec:hyperparams}

To ensure methodological rigor and avoid data leakage, we strictly separate hyperparameter tuning from final evaluation. We randomly partition the 150 RQs into a validation set (50 RQs) for hyperparameter tuning and a test set (100 RQs) for final evaluation. Hyperparameter grid search (Iterations: $\{1,2,3\}$, Revisions: $\{1,2,3\}$) is performed exclusively on the validation set to identify the optimal configuration. The best-performing parameters are then fixed and applied to the held-out test set.

Figure~\ref{fig:hyperparam_results} indicates that the configuration with 3 iterations and 3 revisions achieves the highest scores. This suggests that deeper reasoning cycles with iterative revisions help the multi-agent framework synthesize more complex yet grounded hypotheses.

\section{Conclusion}
\label{sec:conclusion}

We present FIG-MAC, a collaborative multi-agent framework built upon the Fine-grained Inspiration Graph (FIG) for automated scientific hypothesis generation (ASHG). Experiments show that integrating ontology-guided reasoning with role-specialized agents improves both the novelty and reliability of generated hypotheses. Current limitations stem from the domain coverage and completeness of the underlying FIG. Future work will explore scalable cross-domain graph construction, continual graph expansion from emerging literature, and tighter integration with automated experimental evaluation.

\begin{credits}
\subsubsection{\ackname}
This work is supported by the National Natural Science Foundation of China (Grant No.~62272188) and the Fundamental Research Funds for the Central Universities (Grant No.~2662025-XXPY006).

\subsubsection{\discintname}
The authors have no competing interests to declare that are relevant to the content of this article.
\end{credits}

\bibliographystyle{splncs04}
\bibliography{custom}
\end{document}